\pdfoutput=1
\documentclass[11pt, letterpaper, logo, onecolumn, copyright, numbering]{paper_improved}
\usepackage{booktabs}
\usepackage{enumitem}
\usepackage{array}
\usepackage{colortbl}
\usepackage{xcolor}
\usepackage{caption}
\usepackage{tabularray}
\usepackage[utf8]{inputenc} 
\usepackage[T1]{fontenc}    
\usepackage{hyperref}       
\usepackage{url}            
\usepackage{booktabs}       
\usepackage{amsfonts}       
\usepackage{nicefrac}       
\usepackage{microtype}      
\usepackage{xcolor}         
\usepackage{graphicx}

\usepackage{natbib}
\usepackage{verbatim}
\usepackage{inconsolata}
\usepackage{tabularx}
\usepackage{float}
\usepackage{algorithmic}
\usepackage{multirow}
\usepackage{listings}
\usepackage{xcolor}
\usepackage{subfigure}
\usepackage{subcaption}
\usepackage{amsmath}
\usepackage{amssymb}
\usepackage{array}
\usepackage[ruled,linesnumbered]{algorithm2e}
\usepackage[T1]{fontenc}
\usepackage{fancyhdr}

\usepackage[most]{tcolorbox}
\usepackage{alltt}
\usepackage{lineno}
\newtcolorbox{AIbox}[2][]{aibox,title=#2,#1}
\tcbset{
  aibox/.style={
    top=10pt,
    colframe=black,
    colbacktitle=black,
    enhanced,
    center,
    attach boxed title to top left={yshift=-0.1in,xshift=0.15in},
    boxed title style={boxrule=0pt,colframe=white,},
  }
}

\definecolor{lightgray}{rgb}{0.95, 0.95, 0.95}
\definecolor{darkgray}{rgb}{0.4, 0.4, 0.4}
\definecolor{backcolour}{rgb}{0.95,0.95,0.92}
\definecolor{myblue}{rgb}{0.2, 0.4, 0.8} 
\definecolor{mygreen}{rgb}{0.2, 0.6, 0.2} 

\usepackage{amsmath,amssymb,amsfonts}

\let\cite\citep

\title{Marco-o1: Towards Open Reasoning Models for Open-Ended Solutions}

\reportnumber{} 

\author[*,1]{Yu Zhao{$^*$}, Huifeng Yin{$^*$}, Bo Zeng, Hao Wang, Tianqi Shi, Chenyang Lyu, Longyue Wang, Weihua Luo, Kaifu Zhang\\~\\ \bf MarcoPolo Team, Alibaba International Digital Commerce}

\begin{abstract}
Currently OpenAI o1 sparks a surge of interest in the study of large reasoning models (LRM). Building on this momentum, Marco-o1 not only focuses on disciplines with standard answers, such as mathematics, physics, and coding---which are well-suited for reinforcement learning (RL)---but also places greater emphasis on open-ended resolutions. We aim to address the question: ``Can the o1 model effectively generalize to broader domains where clear standards are absent and rewards are challenging to quantify?'' Marco-o1 is powered by Chain-of-Thought (CoT) fine-tuning, Monte Carlo Tree Search (MCTS), reflection mechanisms, and innovative reasoning strategies---optimized for complex real-world problem-solving tasks. The project homepage is: {\url{https://github.com/AIDC-AI/Marco-o1}.}
\end{abstract}

\footnotetext[1]{Equal Contribution.}
\footnotetext[2]{Email: wanglongyue.wly@alibaba-inc.com.}

\begin{document}

\maketitle

\begin{figure}[h]
\centering
\includegraphics[width=0.75\textwidth]{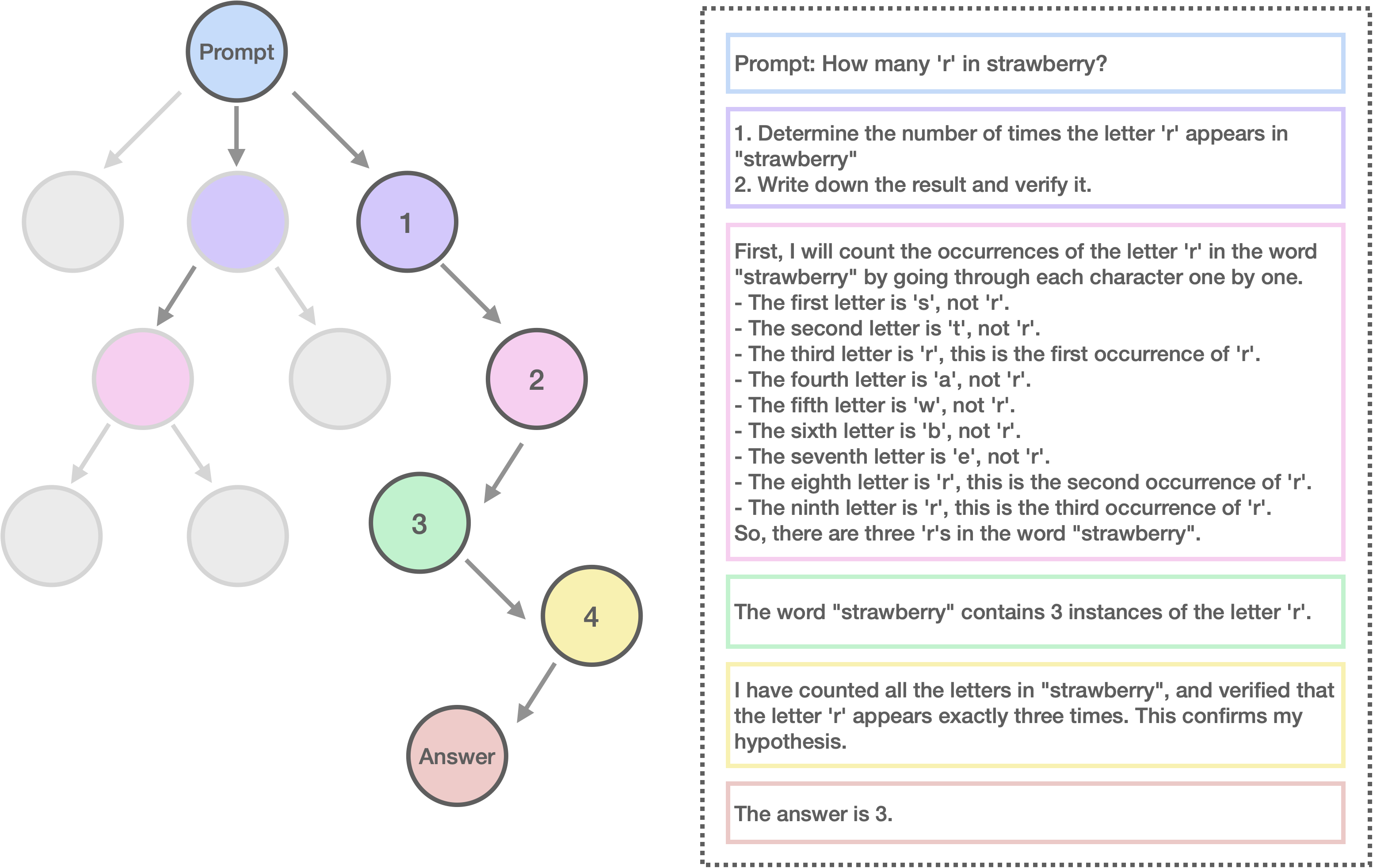}
\caption{A classic question reasoned by our Marco-o1 model: ``How many `r's are in `strawberry'.''}
\label{fig:strawberry}
\end{figure}

\begin{AIbox}{Work in Progress}

\small We would like to emphasize that this research work is inspired by OpenAI's o1 (from which the name is also derived). This work aims to explore potential approaches to shed light on the currently unclear technical roadmap for large reasoning models. Besides, our focus is on open-ended questions, and we have observed interesting phenomena in multilingual applications. However, we must acknowledge that the current model primarily exhibits o1-like reasoning characteristics and its performance still fall short of a fully realized "o1" model. This is not a one-time effort, and we remain committed to continuous optimization and ongoing improvement.

\end{AIbox}

\section{Introduction}

OpenAI recently introduces the groundbreaking o1 model~\cite{openai2024reason, zhong2024evaluation}, renowned for its exceptional reasoning capabilities. This model has demonstrates outstanding performance on platforms such as AIME and CodeForces, surpassing other leading models. Inspired by this success, we aim to push the boundaries of LLMs even further, enhancing their reasoning abilities to tackle complex, real-world challenges.

\begin{figure}[t]
\centering
\includegraphics[width=0.8\textwidth]{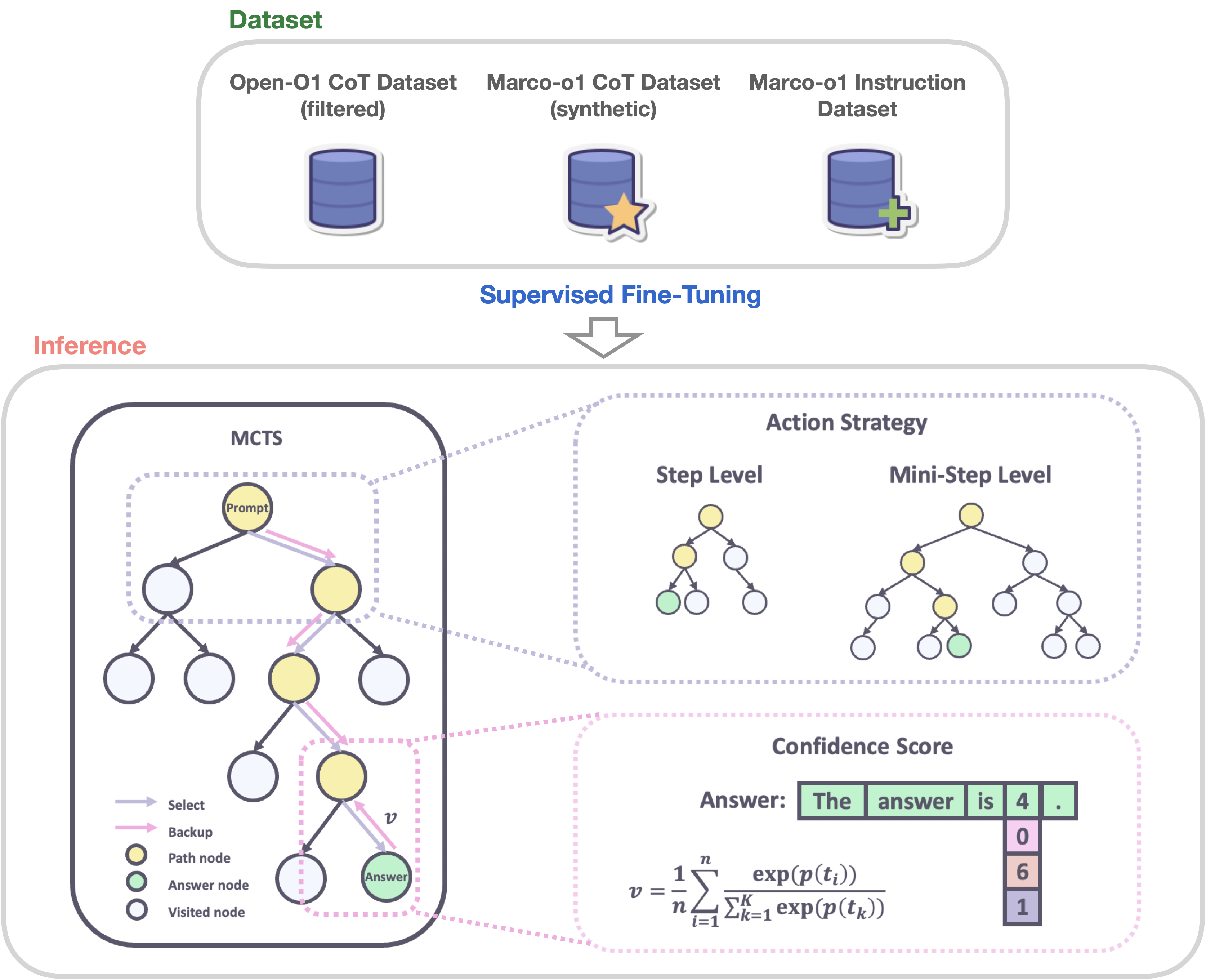}
\caption{The overview of Marco-o1.}
\label{fig:overview}
\end{figure}

Inspired by OpenAI's o1, we aim to explore potential approaches to shed light on the currently unclear technical roadmap for large reasoning models (LRM). Marco-o1 leverages advanced techniques like CoT fine-tuning~\cite{wei2022chain}, MCTS~\cite{wei2022chain, feng2023alphazero, silver2017mastering}, and Reasoning Action Strategies to enhance its reasoning power. As shown in Figure~\ref{fig:overview}, by fine-tuning Qwen2-7B-Instruct~\cite{yang2024qwen2} with a combination of the filtered Open-O1 CoT dataset~\cite{openo1team2024openo1}, Marco-o1 CoT dataset, and Marco-o1 Instruction dataset, Marco-o1 improves its handling of complex tasks. MCTS allows exploration of multiple reasoning paths using confidence scores derived from softmax-applied log probabilities of the top-$k$ alternative tokens, guiding the model to optimal solutions. Moreover, our reasoning action strategy involves varying the granularity of actions within steps and mini-steps to optimize search efficiency and accuracy.

Preliminary experiments demonstrates that our model can exhibit o1-like reasoning characteristics. Furthermore, Marco-o1 achieved accuracy improvements of +6.17\% on the MGSM (English) dataset and +5.60\% on the MGSM (Chinese) dataset, showcasing enhanced reasoning capabilities~\cite{shi2022language}. Additionally, in translation tasks, we demonstrate that Marco-o1 excels in translating slang expressions. For example, the model correctly translates a colloquial expression in Chinese that literally means ``This shoe offers a stepping-on-poop sensation'' to English ``This shoe has a comfortable sole,'' demonstrating its superior grasp of colloquial nuances.
Currently, our {\bf main contributions} are:

\begin{itemize}[leftmargin=*,topsep=0.1em,itemsep=0.1em,parsep=0.1em]
\item \textbf{Fine-Tuning with CoT Data:} We develop \underline{Marco-o1-CoT} by performing full-parameter fine-tuning on the base model using open-source CoT datasets combined with our synthetic data.
\item \textbf{Solution Space Expansion via MCTS:} We integrate LLMs with MCTS (\underline{Marco-o1-MCTS}), using the model's output confidence to guide the search and expand the solution space.
\item \textbf{Reasoning Action Strategy:} We implement novel reasoning action strategies and a reflection mechanism (\underline{Marco-o1-MCTS mini-step}), including exploring different action granularities within the MCTS framework and prompting the model to self-reflect, thereby significantly enhancing the model's ability to solve complex problems.
\item \textbf{Application in Translation Tasks:} We are the first to investigate LRM on \underline{Machine Translation tasks}, exploring inference-time scaling laws in the multilingual and translation domain.
\end{itemize}

\section{Marco Reasoning Datasets}

To enhance the reasoning capabilities of the Marco-o1 model, we employ a Supervised Fine-Tuning (SFT) strategy using a variety of datasets.

\begin{itemize}[leftmargin=*,topsep=0.1em,itemsep=0.1em,parsep=0.1em]
\item  \textbf{Open-O1 CoT Dataset (Filtered)~\cite{openo1team2024openo1}:} We refine the Open-O1 project's CoT Dataset by applying heuristic and quality filtering processes. This enhancement allows the model to adopt structured reasoning patterns effectively.
\item  \textbf{Marco-o1 CoT Dataset (Synthetic):} We generate the Marco-o1 CoT Dataset using MCTS, which helps to formulate complex reasoning pathways, bolstering the model's reasoning capabilities.
\item  \textbf{Marco Instruction Dataset:} Recognizing the critical role of robust instruction-following capabilities in executing complex tasks, we incorporate a set of instruction-following data. This integration ensures the model remains competent across a wide range of tasks, maintaining its general effectiveness while significantly boosting its reasoning flair.
\end{itemize}

\begin{table}[t]
\centering
\begin{tabular}{l r}
\toprule
\textbf{Dataset} & \textbf{Number of Samples} \\
\midrule
Open-O1 CoT Dataset (Filtered)~\cite{openo1team2024openo1} & 45,125 \\
Marco-o1 CoT Dataset (Synthetic) & 10,000 \\
Marco Instruction Dataset & 5,141 \\
\textbf{Total} & \textbf{60,266} \\
\bottomrule
\end{tabular}
\caption{Overview of Marco Reasoning Datasets.}
\end{table}


\section{Solution Space Expansion via MCTS}

We integrate LLMs with MCTS to enhance the reasoning capabilities of our Marco-o1 model:

\begin{itemize}[leftmargin=*,topsep=0.1em,itemsep=0.1em,parsep=0.1em]
\item  \textbf{Nodes as Reasoning States:} In the MCTS framework, each node represents a reasoning state of the problem-solving process.
\item  \textbf{Actions as LLM Outputs:} The possible actions from a node are the outputs generated by the LLM. These outputs represent potential steps or mini-steps in the reasoning chain.
\item  \textbf{Rollout and Reward Calculation:} During the rollout phase, the LLM continues the reasoning process to a terminal state.
\item  \textbf{Guiding MCTS:} This reward score $R$ is used to evaluate and select promising paths within the MCTS, effectively guiding the search towards more confident and reliable reasoning chains.
\end{itemize}

Furthermore, we obtain the value of each state by computing a confidence score. For each token $t_i$ generated during the rollout, we calculate its confidence score by applying the softmax function to its log probability and the log probabilities of the top 5 alternative tokens. This is given by:

\[
c_i = \frac{\exp(p(t_i))}{\sum_{k=1}^{5} \exp(p(t_k))}
\]

\noindent where $c_i$ is the confidence score for the $i^{th}$ token in the rollout. $p(t_i)$ is the log probability of the $i^{th}$ token generated by the LLM. $p(t_k)$ for $k = 1$ to $5$ are the log probabilities of the top 5 predicted tokens at the $i^{th}$ step. $n$ is the total number of tokens in the rollout sequence. This equation ensures that the confidence score reflects the relative probability of the chosen token compared to the top alternatives, effectively normalizing the scores between 0 and 1.

After obtaining the confidence scores for all tokens in the rollout sequence, we compute the average confidence score across all tokens to derive the overall reward score:

\[
v = \frac{1}{n} \sum_{i=1}^{n} c_i
\]

\noindent where $v$ is the overall reward score for the rollout path. This average serves as the reward signal that evaluates the quality of the reasoning path taken during the rollout. A higher $v$ indicates a more confident and likely accurate reasoning path.

By employing this method, we effectively expand the solution space, allowing the model to explore a vast array of reasoning paths and select the most probable ones based on calculated confidence scores.

\section{Reasoning Action Strategy}

\subsection{Action Selection}

We observe that using actions as the granularity for MCTS search is relatively coarse, often causing the model to overlook nuanced reasoning paths crucial for solving complex problems. To address this, we explore different levels of granularity in the MCTS search. Initially, we use steps as the unit of search. To further expand the model's search space and enhance its problem-solving capabilities, we experiment with dividing these steps into smaller units of 64 or 32 tokens, referred to as ``mini-step.'' This finer granularity allows the model to explore reasoning paths in greater detail. While token-level search offers theoretical maximum flexibility and granularity, it is currently impractical due to the significant computational resources required and the challenges associated with designing an effective reward model at this level.

In our experiments, we implement the following strategies within the MCTS framework:

\begin{itemize}[leftmargin=*,topsep=0.1em,itemsep=0.1em,parsep=0.1em]
\item  \textbf{Step as Action:} We allow the model to generate complete reasoning steps as actions. Each MCTS node represents an entire thought or action label. This method enables efficient exploration but may miss finer-grained reasoning paths essential for complex problem-solving.

\item  \textbf{Mini-step as Action:} We use mini-steps of 32 or 64 tokens as actions. This finer granularity expands the solution space and improves the model's ability to navigate complex reasoning tasks by considering more nuanced steps in the search process. By exploring the solution space at this level, the model is better equipped to find correct answers that might be overlooked with larger action units.
\end{itemize}

\subsection{Reflection after Thinking}

We introduce a reflection mechanism by adding the phrase \textit{``Wait! Maybe I made some mistakes! I need to rethink from scratch.''} at the end of each thought process. This prompts the model to self-reflect and reevaluate its reasoning steps. Implementing this reflection yields significant improvements, especially on difficult problems that the original model initially solves incorrectly. With the addition of reflection, approximately half of these challenging problems are answered correctly.

From the self-critic perspective~\cite{valmeekam2023can}, this approach allows the model to act as its own critic, identifying potential errors in its reasoning. By explicitly prompting the model to question its initial conclusions, we encourage it to re-express and refine its thought process. This self-critical mechanism leverages the model's capacity to detect inconsistencies or mistakes in its own output, leading to more accurate and reliable problem-solving~\cite{madaan2024self, li2024guiding, huang2022large}. The reflection step serves as an internal feedback loop, enhancing the model's ability to self-correct without external intervention.

\section{Experiments}

\subsection{Setup}

Based on \textbf{Qwen2-7B-Instruct},\footnote{https://huggingface.co/Qwen/Qwen2-7B-Instruct} we perform SFT using our training data to create \textbf{Marco-o1-CoT}. Besides, we employ Marco-o1-CoT within the framework of MCTS tree search, differentiating by:

\begin{itemize}[leftmargin=*,topsep=0.1em,itemsep=0.1em,parsep=0.1em]
\item  \textbf{Marco-o1-MCTS (step)}: using each inference step as an action (step).
\item  \textbf{Marco-o1-MCTS (mini-step of 64 tokens)}: using a 64-token mini-step as an action (64 tokens).
\item  \textbf{Marco-o1-MCTS (mini-step of 32 tokens)}: using a 32-token mini-step as an action (32 tokens).
\end{itemize}

During testing, each model utilizes a CoT prompt to ensure consistency in reasoning processes. We then test these configurations on the English (En) and Chinese (Zh) subsets of the MGSM dataset.

\begin{table}[t]
\centering
\begin{tabular}{lcc}
\toprule
\textbf{Model} & \textbf{MGSM-En (Acc.)} & \textbf{MGSM-Zh (Acc.)} \\
\midrule
Qwen2-7B-Instruct & 84.00\% & 76.80\% \\
Marco-o1-CoT & 85.60\% & 71.20\% \\
Marco-o1-MCTS (step) & 90.40\% & 80.00\% \\
Marco-o1-MCTS (mini-step of 64 tokens) & 88.40\% & 80.40\% \\
Marco-o1-MCTS (mini-step of 32 tokens) & 87.60\% & 82.40\% \\
\bottomrule
\end{tabular}
\caption{Experimental results on MGSM datasets.}
\label{tab:results}
\end{table}

\begin{figure}[ht]
\centering
\includegraphics[width=0.8\textwidth]{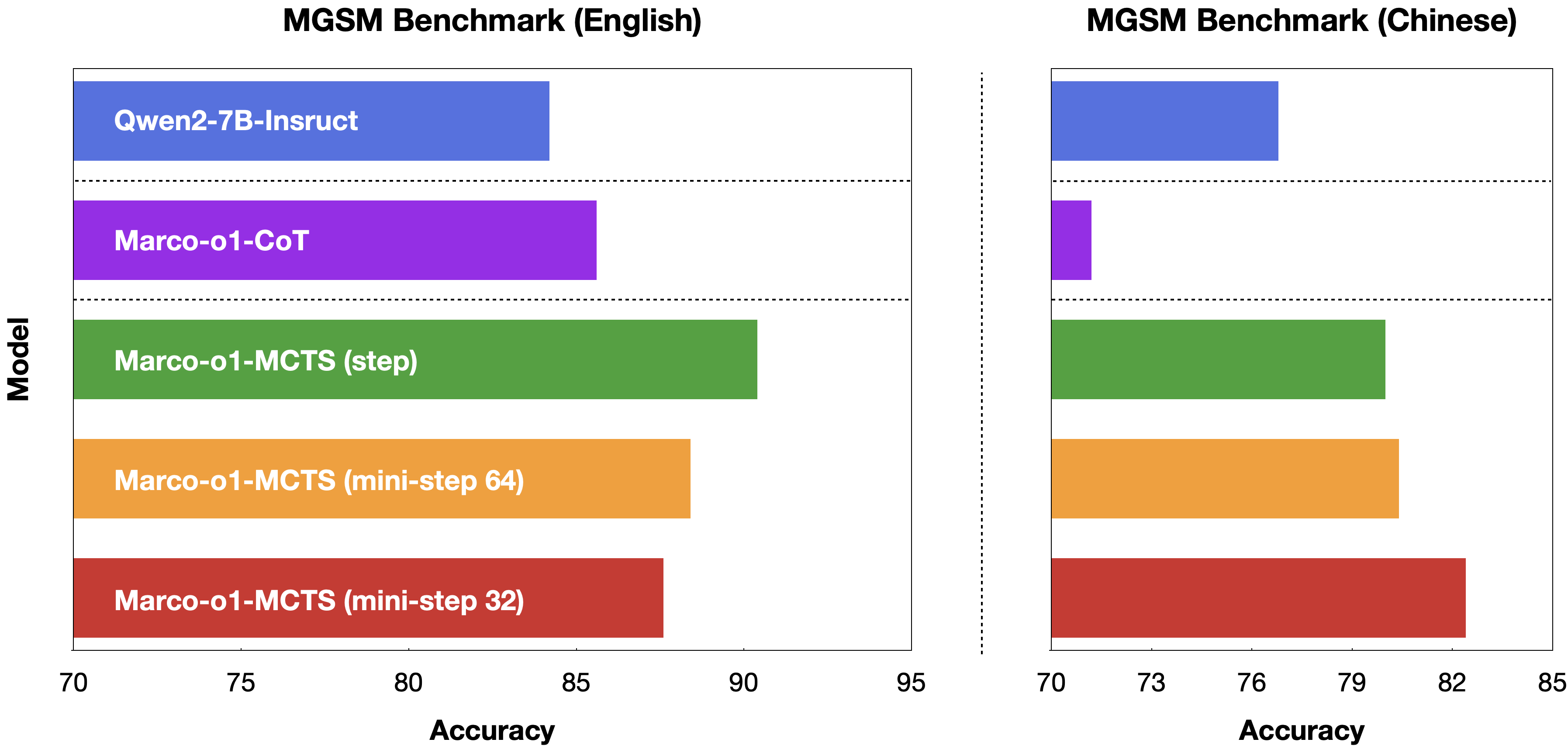}
\caption{The main results of Marco-o1.}
\label{fig:results}
\end{figure}

\begin{table}[t]
\centering
\begin{tabular}{lccc}
\toprule
 & & \textbf{MGSM-En} & \\
\cmidrule(lr){2-4}
\textbf{Model} & \textbf{Test@1 Acc.} & \textbf{Test@8 Acc.} & \textbf{Test@32 Acc.} \\
\midrule
Qwen2-7B-Instruct & 84.00\% & 89.60\% & 96.00\% \\
Marco-o1-CoT & 85.60\% & 97.60\% & 99.20\% \\
Marco-o1-MCTS (step) & 90.40\% & 99.20\% & 99.20\% \\
Marco-o1-MCTS (mini-step of 64 tokens) & 88.40\% & 98.40\% & 99.60\% \\
Marco-o1-MCTS (mini-step of 32 tokens) & 87.60\% & 98.80\% & 99.20\% \\
\midrule
 & & \textbf{MGSM-Zh} & \\
\cmidrule(lr){2-4}
\textbf{Model} & \textbf{Test@1 Acc.} & \textbf{Test@8 Acc.} & \textbf{Test@32 Acc.} \\
\midrule
Qwen2-7B-Instruct & 76.80\% & 80.80\% & 92.40\% \\
Marco-o1-CoT & 71.20\% & 93.60\% & 96.40\% \\
Marco-o1-MCTS (step) & 80.00\% & 93.60\% & 96.00\% \\
Marco-o1-MCTS (mini-step of 64 tokens) & 80.40\% & 92.80\% & 95.20\% \\
Marco-o1-MCTS (mini-step of 32 tokens) & 82.40\% & 93.20\% & 96.80\% \\
\bottomrule
\end{tabular}
\caption{Performance on MGSM Datasets: Test@1, Test@8, and Test@32 Results. Test@N denotes the percentage of problems solved correctly at least once when the model is allowed to make N separate guesses for each problem.}
\label{tab:results2}
\end{table}


\subsection{Main Results}

\begin{figure}[ht]
\centering
\includegraphics[width=\textwidth]{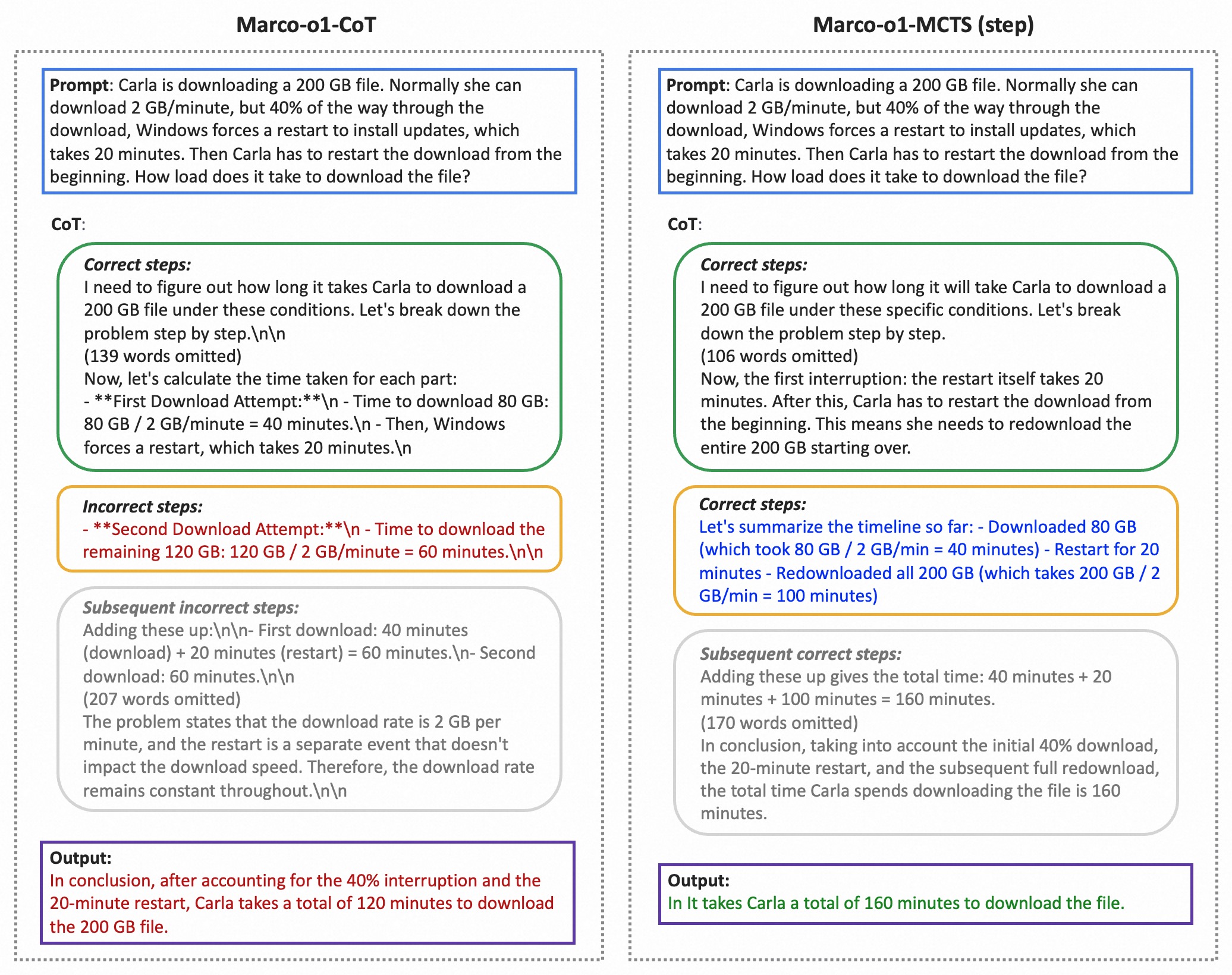}
\caption{MCTS Expands the Solution Space for Correct Answers. Comparison between Marco-o1-CoT (left) and Marco-o1-MCTS (step) (right) on the MGSM dataset. While Marco-o1-CoT fails to provide the correct answer, integrating MCTS with step-level actions allows the model to explore a broader solution space, increasing the likelihood of arriving at the correct solution.}
\label{fig:cot-step}
\end{figure}

In the MGSM-en dataset, Marco-o1-CoT shows an advantage over Qwen2-7B-Instruct, as shown in Figure~\ref{fig:cot-step}, which is expected due to the fine-tuning with English CoT data. In the MGSM-zh dataset, however, Marco-o1-CoT exhibits a decrease in performance compared to Qwen2-7B-Instruct. This decline is attributed to the fact that the CoT data used for fine-tuning was in English, which may not transfer effectively to the Chinese dataset.

\begin{figure}[ht]
\centering
\includegraphics[width=\textwidth]{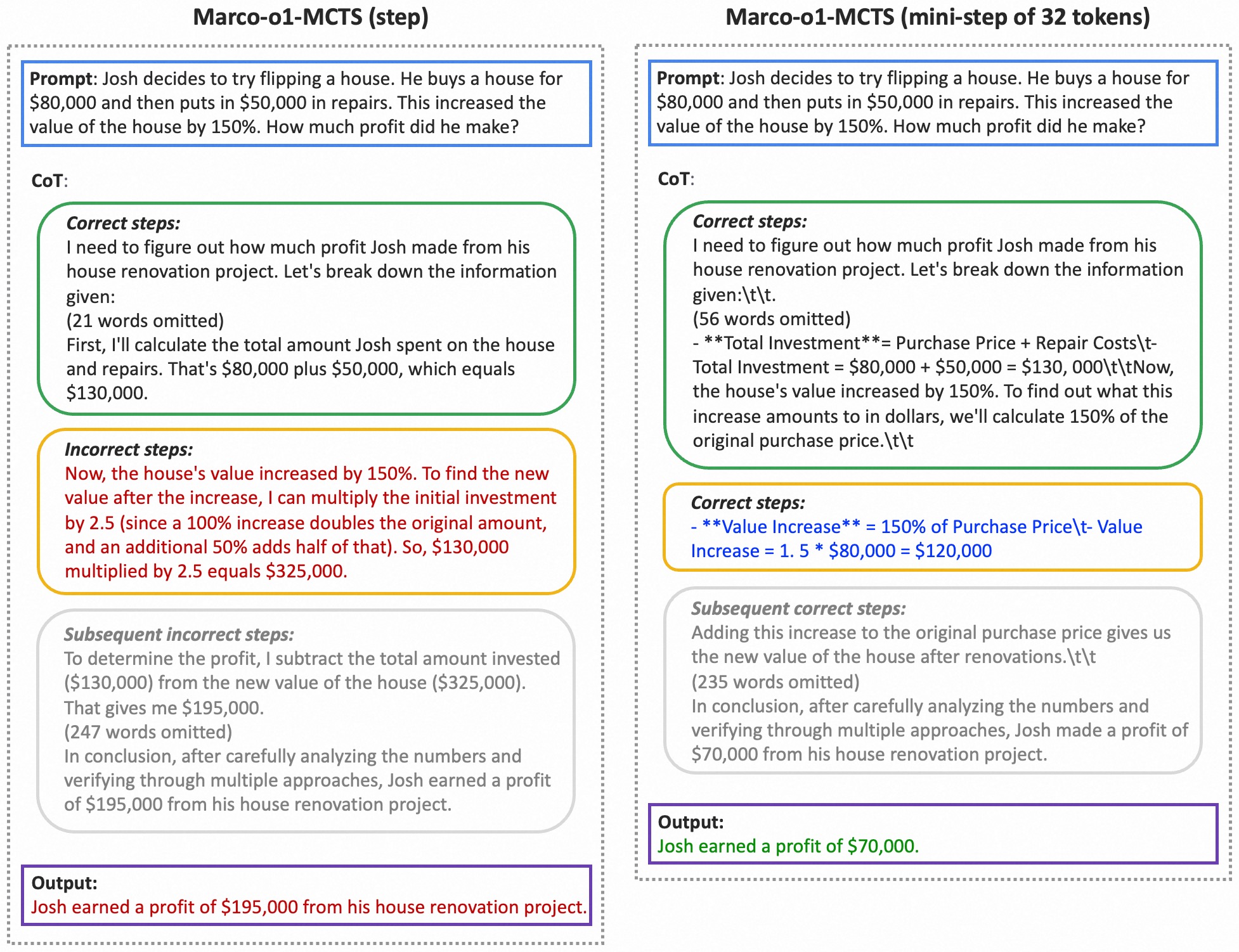}
\caption{Finer Granularity with mini-steps Enhances Problem-Solving. Comparison between Marco-o1-MCTS (step) (left) and Marco-o1-MCTS (mini-step of 32 tokens) (right) on the MGSM dataset. The step-level action strategy does not yield the correct answer, but by using a finer-grained mini-step of 32 tokens, the model successfully navigates the solution space to find the correct answer, demonstrating the effectiveness of increased action granularity.}
\label{fig:step-ministep32}
\end{figure}

\begin{figure}[ht]
\centering
\includegraphics[width=\textwidth]{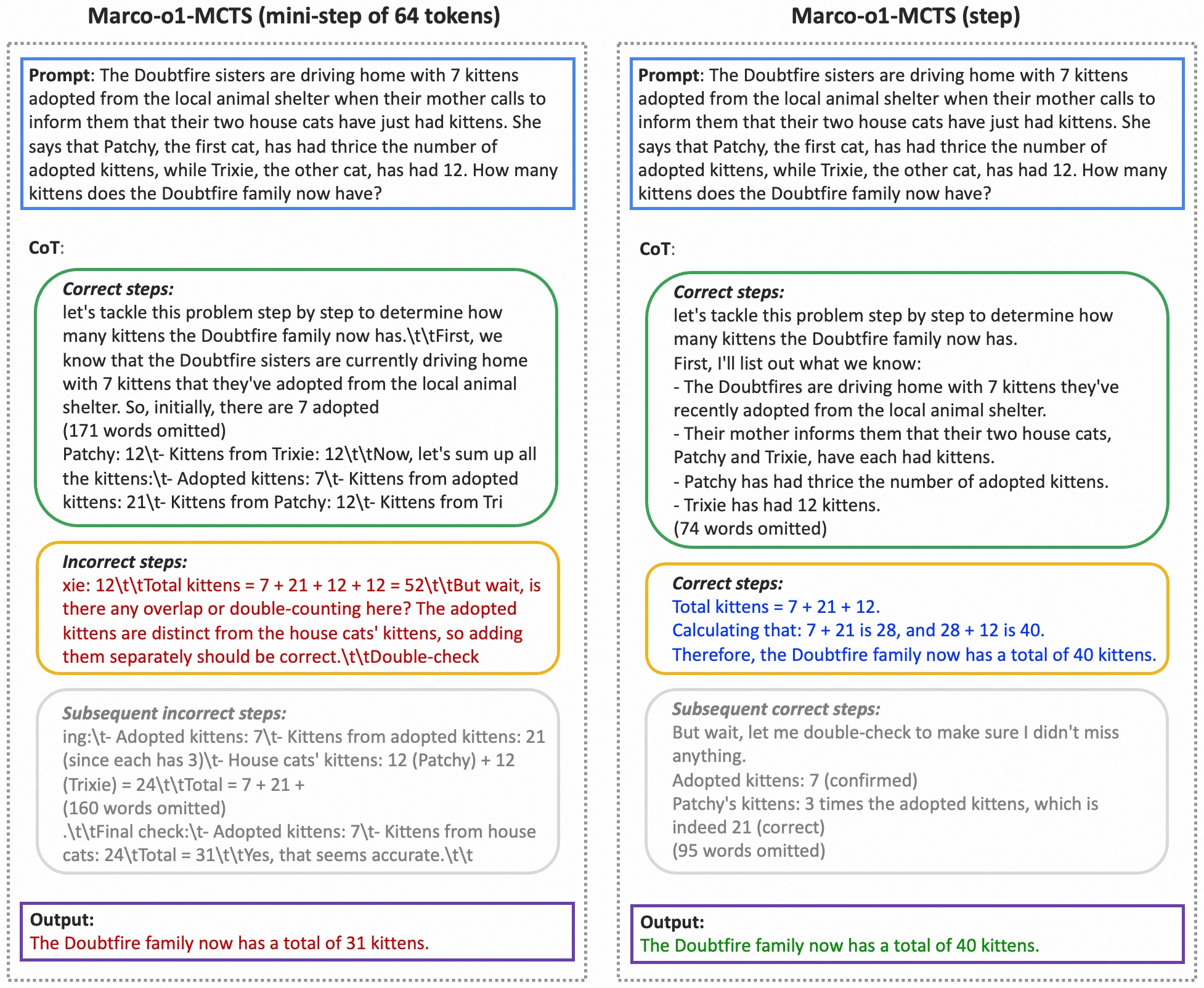}
\caption{Optimal Action Granularity Depends on Problem Complexity. Comparison between Marco-o1-MCTS (mini-step of 64 tokens) (left) and Marco-o1-MCTS (step) (right) on the MGSM dataset. The model with a mini-step of 64 tokens fails to find the correct answer, whereas using step-level actions enabled the model to correctly solve the problem. This highlights that we cannot draw definitive conclusions about which action strategy is superior. We believe that as the reward becomes more accurate, the larger solution space provided by MCTS will demonstrate greater potential.}
\label{fig:ministep64-step}
\end{figure}

As shwon in Table~\ref{tab:results}, The three MCTS-enhanced models demonstrate improvements over Marco-o1-CoT, indicating that incorporating MCTS helps to expand the model's solution space and increase the probability of obtaining correct answers. However, since we use the Confidence Score as the reward, the tree search results exhibit significant randomness. In MGSM-en, the ``step as Action'' strategy performs the best, while in MGSM-zh, the ``mini-step as Action (32)'' strategy yields the highest accuracy. Currently, as shown in Figures~\ref{fig:cot-step}, \ref{fig:step-ministep32}, and \ref{fig:ministep64-step}, we cannot draw definitive conclusions about which action strategy is superior. We believe that as the reward becomes more accurate, the larger solution space provided by MCTS will demonstrate greater potential.

Furthermore, we use Test@N to denote the percentage of problems solved correctly at least once when allowing the model to make N separate guesses for each problem.\cite{cobbe2021trainingverifierssolvemath} As shwon in Table~\ref{tab:results2}, we evaluated solve rates at Test@1, Test@8, and Test@32. The results demonstrate that MCTS shows an advantage with a lower number of separate guesses (Test@1). This reveals the potential of MCTS. In future work, we plan to train the reward model (RM) in conjunction with MCTS to continue optimizing our approach.

An interesting observation arises in Figure~\ref{fig:strawberry}, where Marco-o1 tackles the classic `strawberry' question: ``How many `r's are in `strawberry'.'' While the model provides the correct answer, its reasoning omits explicitly addressing the final letter `y'. Upon reviewing the output, this omission does not appear to be a typo or intentional oversight. In the MCTS search paths, most routes also ignored the `y', though some do include it. This suggests a possibility that the model might be confident enough in its reasoning to skip explicitly addressing the final step, akin to how humans sometimes omit obvious steps when solving problems.

These results demonstrate the effectiveness of our approach in enhancing the reasoning capabilities of the model across different languages and configurations.

\subsection{Case Study on Translation Tasks}

To demonstrate the capabilities of our Marco-o1 model in translation tasks, we conduct a case study comparing its performance with Google Translate on translating colloquial and slang expressions. Our model excels in understanding context and nuances, providing more accurate and natural translations.

\begin{figure}[ht]
\centering
\includegraphics[width=0.8\textwidth]{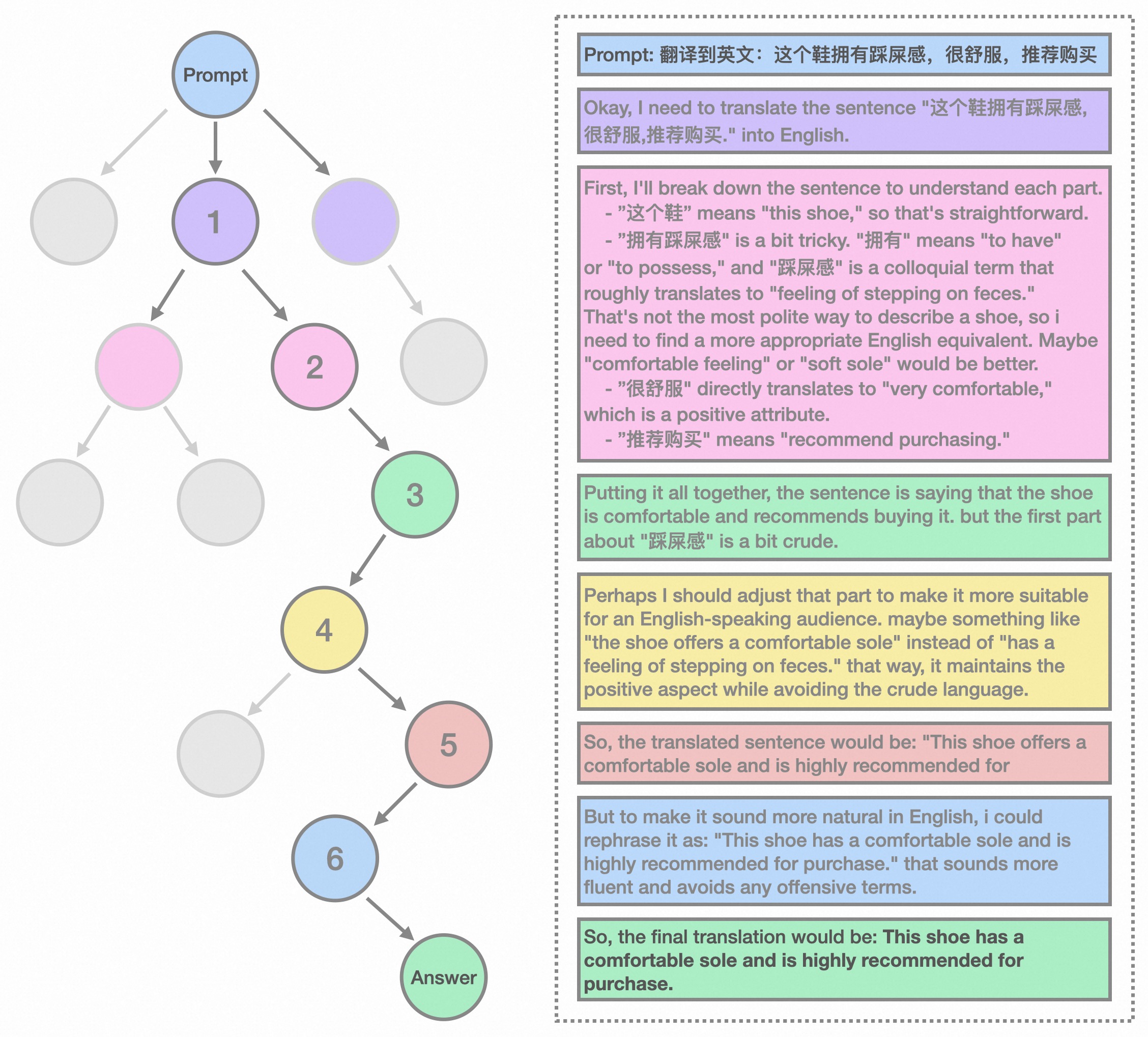}
\caption{Demonstration of translation task using Marco-o1 of a colloquial expression ``This shoe has a comfortable sole and is highly recommended for purchase''.}
\label{fig:translation}
\end{figure}

\begin{figure}[ht]
\centering
\includegraphics[width=0.8\textwidth]{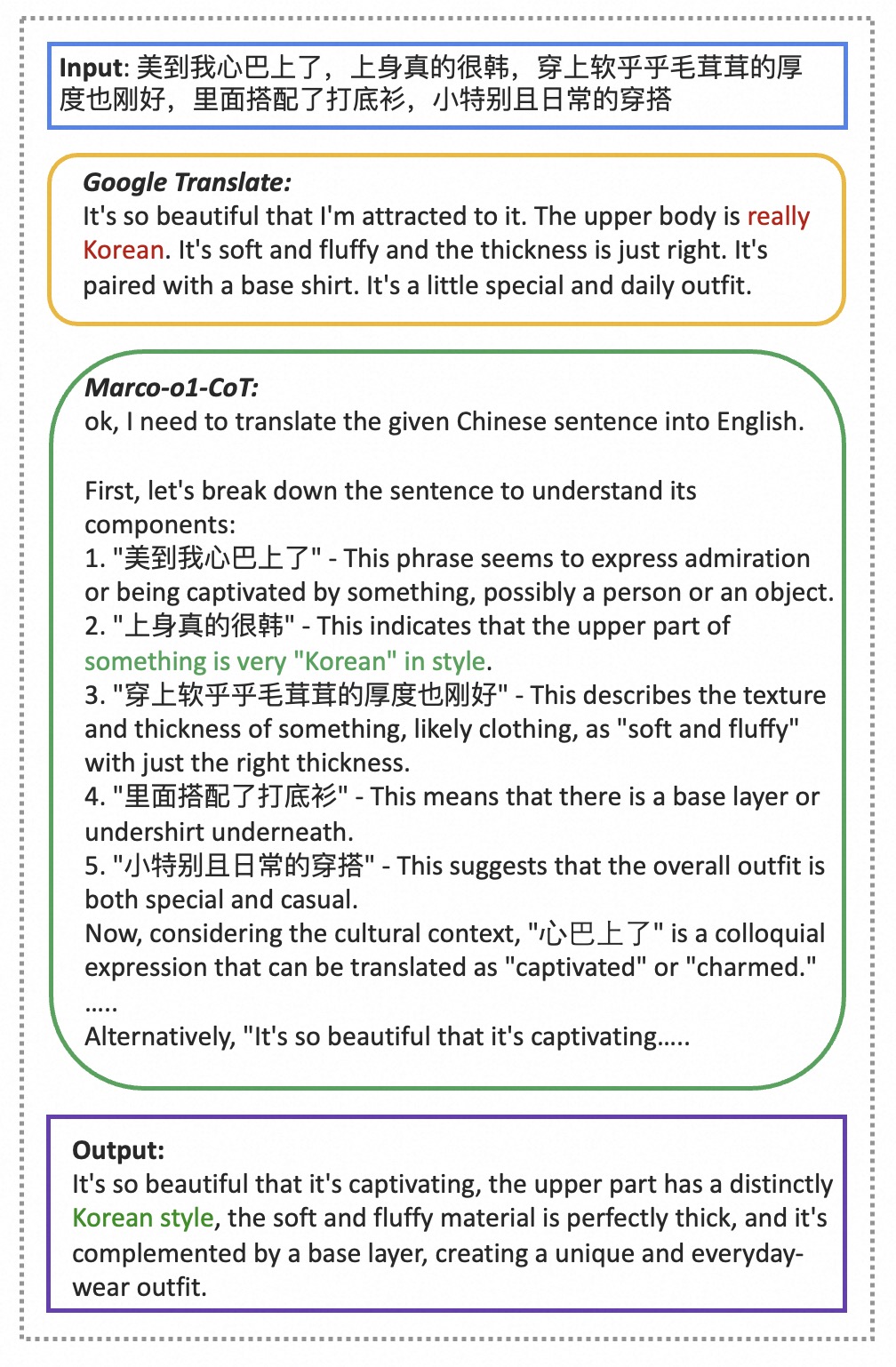}
\caption{Translation comparison of a colloquial expression ``It's so beautiful that it's captivating, the upper part has a distinctly Korean style, the soft and fluffy material is perfectly thick, and it's complemented by a base layer, creating a unique and everyday-wear outfit''.}
\label{fig:translation1}
\end{figure}

\begin{figure}[ht]
\centering
\includegraphics[width=0.8\textwidth]{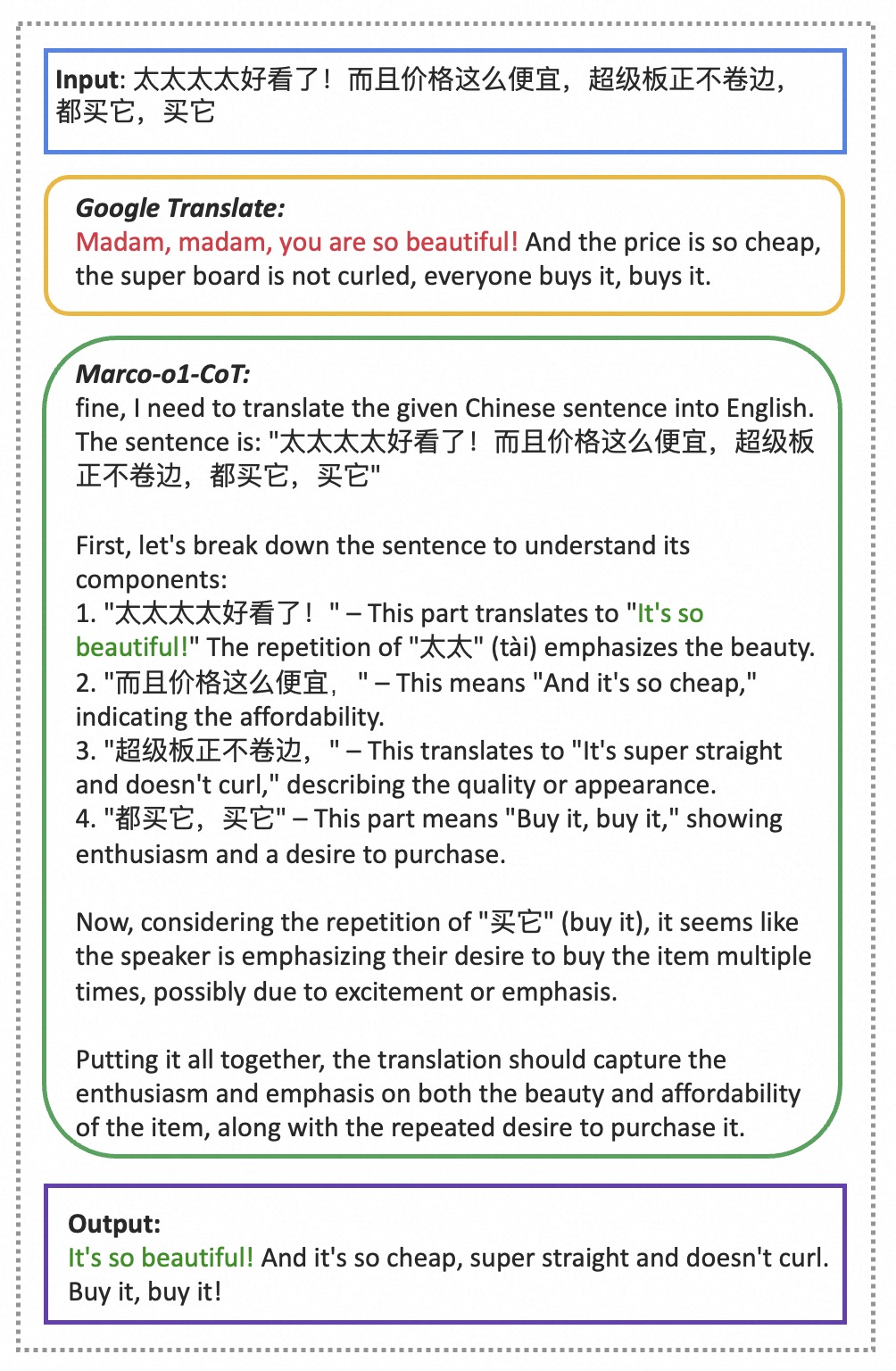}
\caption{Translation comparison of a colloquial expression ``It's so beautiful! And it's so cheap, super straight and doesn't curl. Buy it, buy it!''.}
\label{fig:translation2}
\end{figure}

As shown in Figure~\ref{fig:translation}, ~\ref{fig:translation1}, and ~\ref{fig:translation2}, these cases illustrate the advanced understanding and reasoning capabilities of Marco-o1 in handling complex translation tasks, especially with colloquial and slang language, outperforming standard translation tools like Google Translate.

\section{Conclusions and Future Work}

Our Marco-o1 enhances the reasoning ability by integrating Chain-of-Thought (CoT) fine-tuning, Monte Carlo Tree Search (MCTS), and novel reasoning action strategies. Marco-o1’s integration of MCTS allows for expanded solution spaces, and experimentation with different action granularities (steps and mini-steps) shows the potential of finer search resolutions in enhancing accuracy. Our approach demonstrates significant improvements in reasoning tasks, as well as success in translating complex slang expressions. 

Looking ahead, we aim to refine the reward signal for MCTS through Outcome Reward Modeling (ORM) and Process Reward Modeling (PRM)~\cite{lightman2023let}, which will reduce randomness and further improve performance. Additionally, reinforcement learning techniques are being explored to fine-tune the decision-making processes of Marco-o1, ultimately enhancing its ability to tackle complex real-world tasks.

\bibliography{main}

\begin{thebibliography}{14}
\providecommand{\natexlab}[1]{#1}
\providecommand{\url}[1]{\texttt{#1}}
\expandafter\ifx\csname urlstyle\endcsname\relax
  \providecommand{\doi}[1]{doi: #1}\else
  \providecommand{\doi}{doi: \begingroup \urlstyle{rm}\Url}\fi

\bibitem[Cobbe et~al.(2021)Cobbe, Kosaraju, Bavarian, Chen, Jun, Kaiser, Plappert, Tworek, Hilton, Nakano, Hesse, and Schulman]{cobbe2021trainingverifierssolvemath}
K.~Cobbe, V.~Kosaraju, M.~Bavarian, M.~Chen, H.~Jun, L.~Kaiser, M.~Plappert, J.~Tworek, J.~Hilton, R.~Nakano, C.~Hesse, and J.~Schulman.
\newblock Training verifiers to solve math word problems, 2021.
\newblock URL \url{https://arxiv.org/abs/2110.14168}.

\bibitem[Feng et~al.(2023)Feng, Wan, Wen, McAleer, Wen, Zhang, and Wang]{feng2023alphazero}
X.~Feng, Z.~Wan, M.~Wen, S.~M. McAleer, Y.~Wen, W.~Zhang, and J.~Wang.
\newblock Alphazero-like tree-search can guide large language model decoding and training.
\newblock \emph{arXiv preprint arXiv:2309.17179}, 2023.

\bibitem[Huang et~al.(2022)Huang, Gu, Hou, Wu, Wang, Yu, and Han]{huang2022large}
J.~Huang, S.~S. Gu, L.~Hou, Y.~Wu, X.~Wang, H.~Yu, and J.~Han.
\newblock Large language models can self-improve.
\newblock \emph{arXiv preprint arXiv:2210.11610}, 2022.

\bibitem[Li et~al.(2024)Li, Peng, He, Galley, Gao, and Yan]{li2024guiding}
Z.~Li, B.~Peng, P.~He, M.~Galley, J.~Gao, and X.~Yan.
\newblock Guiding large language models via directional stimulus prompting.
\newblock \emph{Advances in Neural Information Processing Systems}, 36, 2024.

\bibitem[Lightman et~al.(2023)Lightman, Kosaraju, Burda, Edwards, Baker, Lee, Leike, Schulman, Sutskever, and Cobbe]{lightman2023let}
H.~Lightman, V.~Kosaraju, Y.~Burda, H.~Edwards, B.~Baker, T.~Lee, J.~Leike, J.~Schulman, I.~Sutskever, and K.~Cobbe.
\newblock Let's verify step by step.
\newblock \emph{arXiv preprint arXiv:2305.20050}, 2023.

\bibitem[Madaan et~al.(2024)Madaan, Tandon, Gupta, Hallinan, Gao, Wiegreffe, Alon, Dziri, Prabhumoye, Yang, et~al.]{madaan2024self}
A.~Madaan, N.~Tandon, P.~Gupta, S.~Hallinan, L.~Gao, S.~Wiegreffe, U.~Alon, N.~Dziri, S.~Prabhumoye, Y.~Yang, et~al.
\newblock Self-refine: Iterative refinement with self-feedback.
\newblock \emph{Advances in Neural Information Processing Systems}, 36, 2024.

\bibitem[OpenAI(2024)]{openai2024reason}
OpenAI.
\newblock Learning to reason with llms.
\newblock \url{https://openai.com/index/learning-to-reason-with-llms/}, 2024.
\newblock [Accessed 19-09-2024].

\bibitem[{OpenO1 Team}(2024)]{openo1team2024openo1}
{OpenO1 Team}.
\newblock Open-o1.
\newblock \url{https://github.com/Open-Source-O1/Open-O1}, 2024.
\newblock [Accessed 19-11-2024].

\bibitem[Shi et~al.(2022)Shi, Suzgun, Freitag, Wang, Srivats, Vosoughi, Chung, Tay, Ruder, Zhou, et~al.]{shi2022language}
F.~Shi, M.~Suzgun, M.~Freitag, X.~Wang, S.~Srivats, S.~Vosoughi, H.~W. Chung, Y.~Tay, S.~Ruder, D.~Zhou, et~al.
\newblock Language models are multilingual chain-of-thought reasoners.
\newblock \emph{arXiv preprint arXiv:2210.03057}, 2022.

\bibitem[Silver et~al.(2017)Silver, Schrittwieser, Simonyan, Antonoglou, Huang, Guez, Hubert, Baker, Lai, Bolton, et~al.]{silver2017mastering}
D.~Silver, J.~Schrittwieser, K.~Simonyan, I.~Antonoglou, A.~Huang, A.~Guez, T.~Hubert, L.~Baker, M.~Lai, A.~Bolton, et~al.
\newblock Mastering the game of go without human knowledge.
\newblock \emph{nature}, 550\penalty0 (7676):\penalty0 354--359, 2017.

\bibitem[Valmeekam et~al.(2023)Valmeekam, Marquez, and Kambhampati]{valmeekam2023can}
K.~Valmeekam, M.~Marquez, and S.~Kambhampati.
\newblock Can large language models really improve by self-critiquing their own plans?
\newblock \emph{arXiv preprint arXiv:2310.08118}, 2023.

\bibitem[Wei et~al.(2022)Wei, Wang, Schuurmans, Bosma, Xia, Chi, Le, Zhou, et~al.]{wei2022chain}
J.~Wei, X.~Wang, D.~Schuurmans, M.~Bosma, F.~Xia, E.~Chi, Q.~V. Le, D.~Zhou, et~al.
\newblock Chain-of-thought prompting elicits reasoning in large language models.
\newblock \emph{Advances in neural information processing systems}, 35:\penalty0 24824--24837, 2022.

\bibitem[Yang et~al.(2024)Yang, Yang, Hui, Zheng, Yu, Zhou, Li, Li, Liu, Huang, et~al.]{yang2024qwen2}
A.~Yang, B.~Yang, B.~Hui, B.~Zheng, B.~Yu, C.~Zhou, C.~Li, C.~Li, D.~Liu, F.~Huang, et~al.
\newblock Qwen2 technical report.
\newblock \emph{arXiv preprint arXiv:2407.10671}, 2024.

\bibitem[Zhong et~al.(2024)Zhong, Liu, Pan, Zhang, Zhou, Liang, Wu, Lyu, Shu, Yu, et~al.]{zhong2024evaluation}
T.~Zhong, Z.~Liu, Y.~Pan, Y.~Zhang, Y.~Zhou, S.~Liang, Z.~Wu, Y.~Lyu, P.~Shu, X.~Yu, et~al.
\newblock Evaluation of openai o1: Opportunities and challenges of agi.
\newblock \emph{arXiv preprint arXiv:2409.18486}, 2024.

\end{thebibliography}

\appendix

\end{document}